\documentclass[11pt, draftcls, onecolumn]{IEEEtran}     %manuscript.pdf
\ifCLASSINFOpdf
   \usepackage[pdftex]{graphicx}
   \DeclareGraphicsExtensions{.pdf,.jpeg,.png}
\else
\fi

\usepackage{multicol}

\begin{document}
\title{A statistical learning approach to color demosaicing}

\author{Jacob H.~Oaknin ~\IEEEmembership{} % <-this % stops a space
\thanks{}} % <-this % stops a space


\markboth{}{}
%\markboth{Journal of \LaTeX\ Class Files,~Vol.~6, No.~1, January~2007}%
%{Shell \MakeLowercase{\textit{et al.}}: Bare Demo of IEEEtran.cls for Journals}



\maketitle

\begin{abstract}
%\boldmath
A statistical learning/inference framework for color demosaicing is presented.
We start with simplistic assumptions about color constancy, and recast color demosaicing
as a blind linear inverse problem: color parameterizes the unknown
kernel,
while brightness takes on the role of a latent variable.
An expectation-maximization algorithm naturally suggests itself for the
estimation of them both.
Then, as we gradually broaden the family of hypothesis where color is learned,
we let our demosaicing behave adaptively, in a manner that reflects our prior
knowledge about the statistics of color images. We show that we can incorporate
realistic, learned priors without essentially changing the complexity of
the simple expectation-maximization algorithm we started with.

\end{abstract}


% \fi
\IEEEpeerreviewmaketitle

\section{Introduction}
\IEEEPARstart{C}{}olor demosaicing purports to regenerate a full color image from data collected by a single array of photocells each of whom is sensitive to only one out of three color separations ({\em red/green/blue}). Such an objective may be attained under the premise that the information contained in the pictured scene doesn't exceed the capacity of the acquisition channel \cite{Shannon:1948}, {\em i.e.} of strong statistical dependencies among the three color 
separations. Understanding such dependencies, and properly exploiting them, is the key to success in the demosaicing endeavor.\\

Most evident is the fact that, away from object boundaries, chrominance\footnote{that is, the point-wise ratio between color separations. We will often refer to it just as color} seems to vary only over rather long scales. All modern demosaicing algorithms exploit this, though, usually, in a rather heuristic fashion. For instance, {\em smooth hue transition} \cite{Cok:1986} uses a bilinear interpolation to estimate, first, the green channel, and then the red-to-green and blue-to-green ratios.  Interpolated red-to-green and blue-to-green differences, instead
of ratios, have been used in \cite{Adams:1996}. Ellaborate versions of these ideas have 
been proposed in \cite{Kimmel:1999}, \cite{Wu:1997}, \cite{Lu:2003}, \cite{Hibbard:1995}, \cite{Laroche:1994}, \cite{Hamilton:1997} and \cite{Chang:1999} among others. 
Typically, all these algorithms have been tested 
on scanned analogical pictures (for a survey see \cite{Li:2008}), for which the
true values of the three color separations are known. 
It is far from obvious, though, why the heuristics that performs optimally on such tests would still do so on digitally captured data, whose statistics are likely different\footnote{it may even change among digital cameras. Some of them position anti-aliasing filters in front
of the sensors; others don't. Even more, the spectral response of the sensors
may also vary among cameras. The spectral content of the illuminant is also 
subject to change}. More importantly, 
performance on the very same data set on which an algorithm has been
trained may have been achieved at the expense of the algorithm's ability
to generalize, {\em i.e.} of performing well on so far unseen data \cite{Bousquet:2004}.\\ 

Bayesian theory provides a framework for the integration of 
learned statistical models
into robust estimates. The first approach of this kind to color demosaicing was proposed by Brainard \cite{Brainard:1994} \cite{Brainard:1997},
who cast the problem as a statistical decision one: given the responses of interleaved subarrays of pixels and a model of the
statistical distribution of color images, what image estimate minimizes the expected reconstruction error?. This question, borrowed from his paper \cite{Brainard:1994},
 highlights the two main difficulties
a Bayesian approach must overcome: 
a) learning a prior model for the
statistical distribution of color images; b) figuring out an inference algorithm that
leverages this knowledge, together with the observed data, into  
optimal estimates. Somehow implicit here is an agreed upon meaning for what is
to be understood as optimality, {\em i.e.} for how reconstruction errors are to be measured. The standard choice
in Bayesian theory is mean squared error (MSE), and the corresponding 
estimator have come to be known as Bayes estimate. To be clear, a Bayes 
estimate
guarantees that whatever prior knowledge we may have is put to optimal use, in a MSE sense. Obviously, its performance will still be
limited by the accuracy of the model it exploits, as priors are
 liable to bias the estimates.\\ 

The simplest choice of prior, beside the obvious non-informative one, is a
multivariate normal. This was the choice Brainad made in his pioneering
work \cite{Brainard:1994}.
It has the huge advantage that a vast mathematical apparatus becomes inmmediately available both for learning and inference \cite{Fukunaga:1990} \cite{Shiryaev:1996}.
In addition, the resulting Bayes estimate is linear \cite{Taubman:2000}, and hence of low computational complexity. 
Its main shortcoming, on the other hand, is that stationary gaussian
distributions cannot capture the presence of singularities, like edges, that are ubiquitous in natural images.\\

Modeling the statistics of images beyond second-order is a tremendous
challenge, given the large number of degrees of freedom involved.
The task is somehow made easier by the consideration of some symmetry 
constraints: it may be argued that images, as stochastic processes,
 are spatially uniform, isotropic,
scale-invariant, and local. This last property is captured by the so called 
Markov random fields (MRF), and, for this reason, they have
become essential tools in modeling the prior statistics of natural images.
Regarding gray-scale ones, efforts to model their statistics seem to 
have coalesced in the 
field of experts framework proposed by Roth {\em et al} \cite{Roth:2005}: a model built upon an 
overcomplete set of spatially-localized, band-pass filters with 
t-student distributions.
Efforts to learn a corresponding set of filters for color images, and 
exploit them for color demosaicing, have been
 reported in \cite{Mairal:2008}.
Hel-Or have built a gaussian
prior based on a steerable wavelet representation \cite{Hel-Or:2002}.\\

Inference with MRF is not an easy task either. Usually, Bayes estimates are
replaced with maximum-a-posteriori (MAP) ones. The resulting optimization
problem turns often to be non-convex, and sophisticated techniques,
like simulated annealing, are needed to tackle them.
This is the strategy adopted by Mukherjee {\em et al} in \cite{Mukherjee:2001}, following the 
pioneering work of Geman and Geman \cite{Geman:1984}. Alternatively, whenever appropriate discretization is 
possible, loopy belief propagation \cite{Kschischang:2001} 
\cite{Yedidia:2005} could be employed.\\

It is the purpose of this paper to develop a scheme for demosaicing that can
incorporate realistic, learned prior knowledge, yet is amenable to efficient
computation. The basic tenets of our approach are the following:

realizing that color is constant over large patches, we regard it as a 
classical parameter. This allows us to recast color demosaicing as a blind linear inverse problem,
with color parameterizing the unknown kernel, and brightness taking on the role of a latent variable.  
Our aim is to learn
the former in an unsupervised manner, then infer the latter from the data.
Such job will be carried out by an expectation-maximization algorithm \cite{Dempster:1977}.
The main obstacle we face is, of course, figuring out a suitable regularizer.  
In this respect we make our second basic assumption, namely, color is 
modeled as a pairwise Markov random field whose neighborhood system
encompasses all non-local color correlations as well as those between color and brightness. As
we will see, this 
basically amounts
to claiming that the size of the local region over which color is expected
to remain constant, and its shape, can be predicted from
the local conditions of illumination and the color of the surrounding background.
A regression model for such functional dependence is
to be learned in advance in a supervised manner. In this respect, our scheme may be seen as providing a context for 
efforts initiated in \cite{Go:2000} \cite{Kapah:2000}, where neural networks were used to learn 
demosaicing interpolators as regressors.\\

\begin{figure}[!ht]
\centering
\includegraphics[width=8cm, height=4.5cm, angle=0]{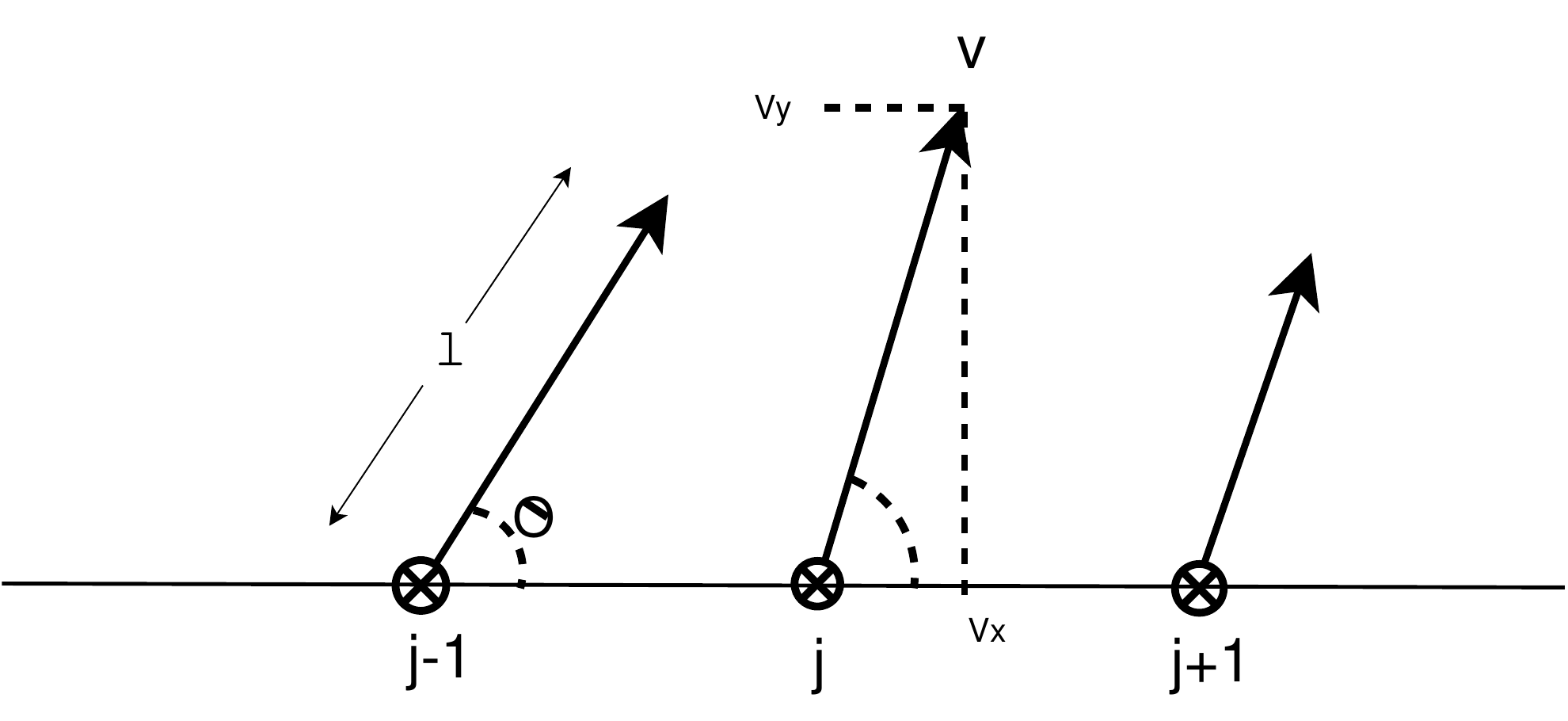}
\caption {{\em Toy problem}. A $1D$ array of pixels, each of whom carries two degrees of freedom,$(v_x, v_y)$. They are jointly represented by vector ${\bf v}$. Color is represented by $\theta$; $l = |{\bf v}|$ represents brightness. At even positions, $j=2k$, only $v_x$ is observed; at odd ones, $j = 2k+1$, only $v_y$} \label{ToyProblem}
\end{figure}

For the sake of clarity we will present our ideas in the context of the simplified {\em toy problem} depicted in  Fig \ref{ToyProblem}  \cite{Kimmel:1999}:
a $1D$ array of pixels each of whom carries only two degrees of freedom, say {\em red} and {\em green}, instead of the usual three separations. They are represented by $\{(v_{x, j}, v_{y, j}), j \in {\textrm Z} \}$, the projection of vectors $\{ {\bf v}_j, j \in {\textrm Z} \}$ on each of the two coordinate axes. The ratio between them, as measured by angles $\{ \theta_j,j \in {\textrm Z} \}$, represents chrominance in this simplified world; brightness is represented by the modulus $\{ l_j  = | {\bf v}_j |, j \in {\textrm Z} \}$. Both projections are non-negative, and so $\theta_j \in \left [ 0, \pi/2 \right ], \forall j$. At even numbered pixels, only $v_x  \, ({\textrm red}) $ is observed; at odd ones, only $v_y \, ({\textrm green})$. All observations are contaminated by white Gaussian noise. We are asked to reconstruct a full vector {\bf v} at each position.\\

The paper is organized as follows: in section {\bf II} we lay down the framework in which our ideas are implemented. We do it in the context of the aforementioned {\em toy problem}.
In section {\bf III} we turn back to our original demosaicing task, and touch on some issues left behind in the previous discussion.
In section {\bf IV} we present the results of a comparative study   
on the performances of a well established, state-of-the-art demosaicing
 algorithm, and of ours. 
We close with a summary of our findings and
directions for future work. 

\section{An Expectation-Maximization estimation of color}

Let's recall our {\em toy problem}, and introduce some notation: a $1D$ array of pixels each of whom carries two degrees of freedom, ${\bf v} = \{ {\bf v}_j = (v_{x, j}, v_{y, j}) ,  \, j \in  {\textrm Z}  \}$. Our observations, ${\bf y} = \{ y_j, \, j \in  {\textrm Z} \}$, comprise a noisy measurement of the 
{\em red} separation at even numbered pixels, and of the {\em green} one, at odd ones: 
\begin{eqnarray} \label{Observations}
y_j =  \left \{ 
\begin{array}{ccc}
v_{x,j} + \epsilon_j & , & {\textrm for} \,\, j = 2k  \\
v_{y, j} + \epsilon_j & , & {\textrm for} \,\, j = 2k+1 
\end{array}
\right . 
\end{eqnarray}
We are asked to reconstruct a full vector {\bf v} at every position. Instead
of cartesian coordinates, we'll be using a polar representation of the pair of degrees of freedom $(v_x, v_y)$: $v_{x, j} = l_j \cos \theta_j$, and $v_{y, j} = l_j \sin \theta_j$. Angles, ${\bf \theta} = \{ \theta_j, j \in {\textrm Z}\}$, as labels of the set of points whose
coordinates
share a common ratio, are proxies for color. Modules, ${\bf l} = \{ l_j, j \in {\textrm Z}\}$, correspond
to brightness. To keep
the notation as neat as possible, we will use the name of a magnitude with no
site label to mean the set of all values of that magnitude in the chain. For
instance, ${\bf \theta}, {\bf l}$ will stand, respectively, for the sets
${\bf \theta} = \{ \theta_j, j \in {\textrm Z}\}$ and ${\bf l} = \{ l_j, j \in {\textrm Z}\}$.\\

Our problem is an ill-posed one, but the following simple observation will shed some light about how we main attempt to regularize it. Let's assume for a moment that our measurements are noise-free, and that color, though unknown, is constant, {\em i.e.} $\theta_j = \theta_0, \forall j$. Under such circumstances it would suffice to require that brightness has no structure whatsoever in the highest frequency mode to turn our problem into a well-posed one. Let's see how 
\begin{eqnarray} \label{HardModel}
\tilde{l}(\pi) = \sum_j e^{i \pi j} l_j = \sum_k l_{2k} - \sum_k l_{2k+1} = 0 
\end{eqnarray} 
where $\tilde{\bf l}$ stands for the discrete Fourier transform  of $\bf l$. In the absence of noise
\begin{eqnarray} \label{NoNoise}
y_j =  \left \{ 
\begin{array}{ccc}
l_j \, \cos \theta_0 & , & {\textrm for} \,\, j = 2k  \\
 l_j \, \sin \theta_0 & , & {\textrm for} \,\, j = 2k+1 
\end{array}
\right . 
\end{eqnarray}
and so, we would conclude
\begin{eqnarray} \label{HardModelEstimate}
\hat{\theta}_0 = \arctan{\frac{\sum_k y_{2k+1}}{\sum_k y_{2k}}} 
\end{eqnarray}
Estimates for ${\bf v}$ are immediately read off from (\ref{NoNoise}) and (\ref{HardModelEstimate}).\\

We now proceed to drop the strong assumptions we have just made. For the sake of clarity we will do it in steps: 
\begin{itemize}
\item [1.] First we drop our requirement that measurements be noiseless, but still assume that color is constant. We replace (\ref{HardModel}) with a realistic {\em prior} for brightness, and (\ref{HardModelEstimate}) with a maximum likelihood estimate.

\item [2.] Next, we relax the assumption that color is constant, and demand from it to be just piece-wise constant. Ideas from robust statistics will be incorporated into a pairwise Markov random field to model this behavior. For the time being the graph structure of the Markov field will be assumed known.

\item [3.] We enrich our color model by letting the graph structure of the Markov field become itself an unobserved variable. As it turns out, what we need is a regression model for the strength of the links between neighbors as a function of the background's color. This is to be learned off-line in a supervised manner.

\item [4.] At last we turn our attention to color-brightness statistical dependencies. As it happens, they can be captured by just letting the regression model mentioned above be a function of both color and brightness. 
\end{itemize}

\subsection{Constant color}

Assuming color is constant, we rewrite (\ref{Observations}) as
\begin{eqnarray} \label{FactorAnalysis}
{\bf y} = \mathcal{D}(\theta_0) \, {\bf l}  + {\bf \epsilon} 
\end{eqnarray}
where $\mathcal{D}(\theta_0) = {\textrm diag}(\dots, cos \theta_0, sin \theta_0, \dots)$ is a diagonal square matrix  with entries $\cos \theta_0$ at even rows, and $\sin \theta_0$ at odd ones. Noise is assumed to be stationary, white and gaussian, ${\bf \epsilon} \sim {\mathcal N}(0; \sigma^2 {\bf I}) $. General covariance matrices could be easily accommodated in our formalism, but these considerations are not within the scope of our interest. Anyhow, we do need to have the photosensors behavior characterized beforehand, so that the covariance matrix of measurements is known in advance.\\

Eq. (\ref{FactorAnalysis}) tries to emphasize one important insight into the problem of color demosaicing, namely, that, because color is almost constant over large patches, we may expect its posterior distribution given the data to be strongly peaked around its mode, and hence regard it as a classical parameter. Optimal estimates for the magnitudes of interest can then be approximated as follows
\begin{eqnarray} \label{OptimalEstimate}
\hat{v}_{x, j} & = & {\textrm E}\left [ v_{x, j} | {\bf y} \right ]   \\
& = & \int d\theta_0 \, d{\bf l} \, v_{x, j}  \, \rho(\theta_0, {\bf l}  |  {\bf y})  \nonumber \\
& = & \int d\theta_0  \, d{\bf l} \,\,  l_j \cos \theta_0 \, \rho({\bf l}  |  {\bf y}, \theta_0)  \, \rho(\theta_0 | {\bf y}) \nonumber \\
& \approx &  {\textrm E}\left [ l_j | {\bf y},  \theta_0^{ML} \right ] \cos \theta_0^{ML}  \nonumber
\\ \nonumber 
\hat{v}_{y, j} & \approx & {\textrm E}\left [ l_j | {\bf y},  \theta_0^{ML} \right ] \sin \theta_0^{ML}  \nonumber
\end{eqnarray}
where 
\begin{eqnarray} \label{ML}
\theta_0^{ML} = {\textrm arg \,\, max \,\, } \rho(\theta_0 | {\bf y}) 
\end{eqnarray}
So, we are tasked with computing $\theta_0^{ML}$, and ${\textrm E}\left [ {\bf l} \, | {\bf y},  \theta_0^{ML} \right ] $. To proceed we need to make explicit our model for the joint distribution of the relevant magnitudes. In making this choice we strive to keep a balance between computational tractability and fidelity. At this stage, we content ourselves with modeling color and brightness as marginally independent. 
 \begin{eqnarray}
 \rho({\bf y}, {\bf l}, \theta_0) = \frac{2}{\pi} \rho({\bf l}) \,\, \rho({\bf y} | {\bf l}, \theta_0)
\nonumber
\end{eqnarray}
Likelihood  and brightness {\em prior}s are respectively given by
\begin{eqnarray}
\rho({\bf y} | {\bf l}, \theta_0)  & = &{\mathcal N}({\bf y} ; {\mathcal D}(\theta_0)  {\bf l} , \, \sigma^2 {\bf I})  \nonumber 
 \end{eqnarray}
\begin{eqnarray}  \label{GaussianPrior}  
\rho({\bf l})  & = & {\mathcal N}({\bf l} ; 0, \, {\bf \Sigma})  
\end{eqnarray}
As usual, ${\mathcal N}(\cdot)$ stands for a normal distribution of its first argument with average and covariance matrix specified in the second and third ones.\\

(\ref{GaussianPrior}) captures only linear correlations. Extensive literature has been devoted to the characterization of the higher-order statistics of natural gray scale images (see for instance \cite{Roth:2005}), and it seems only befitting to incorporate those ideas into $\rho({\bf l})$. Doing so would certainly
yield quality gains to our demosaicing, particularly at high ISOs and low signal-to-noise ratios. In this paper, however, we content ourselves with 
pointing the issue out, but will not pursue it any further.
A second
comment regarding model (\ref{GaussianPrior}) is in place here:
it has been pointed out that natural gray scale images are invariant under scale transforms, as well as stationary and isotropic \cite{Ruderman:1994}. These symmetry requirements impose a power law on its power spectrum \cite{Taubman:2000}
\begin{eqnarray} \label{BrightnessCovariance}
{\bf {\mathcal F} \Sigma {\mathcal F}^{-1}} = {\textrm diag}(\epsilon_0 \, |\omega|^{-\nu}) \nonumber
\end{eqnarray}
${\bf {\mathcal F}}$ stands for the discrete Fourier transform, and $\omega$ is a frequency label.\\

All the ingredients are now in place, and we proceed to compute $\theta_0^{ML}$. Expectation-Maximization  \cite{Dempster:1977} carries out this program by iteratively optimizing a subrogate function
\begin{eqnarray}
\theta_0^{ML} & = & \lim_{t \rightarrow \infty} \theta_0^{(t)} \nonumber \\
\theta_0^{(t+1)} & \equiv & {\textrm arg \,\, max}_{\theta} \, L(\theta ; \theta_0^{(t)}) \nonumber
\end{eqnarray}  
with $L(\theta ; \theta_0^{(t)})$ defined as
\begin{eqnarray}
L(\theta ; \theta_0^{(t)}) \equiv {\textrm E}\left [ {\textrm log} \rho( {\bf y}, {\bf l},  \theta) \big{|} {\bf y}, \theta_0^{(t)} \right ] \nonumber
\end{eqnarray}
In our problem, $L(\theta ; \theta_0^{(t)})$ takes a simple form
\begin{eqnarray}
\sigma^2 L(\theta ; \theta_0^{(t)}) & = & L_0 + P_e \, \cos \theta + P_o \, \sin \theta \\
& & -\frac{1}{2} \{ \Delta_e^2 \, \cos^2\theta + \Delta_o^2 \, \sin^2\theta \} \nonumber
\end{eqnarray}
where $L_0$  doesn't depend on $\theta$, and 
\begin{eqnarray}
P_e({\bf y}, \theta_0^{(t)}) & \equiv & \sum_k y_{2k} {\textrm E}\left [ l_{2k} \big{|} {\bf y}, \theta_0^{(t)} \right ] \nonumber \\
P_o({\bf y}, \theta_0^{(t)}) & \equiv & \sum_k y_{2k+1} {\textrm E}\left [ l_{2k+1} \big{|} {\bf y}, \theta_0^{(t)} \right ] \nonumber \\
\Delta_e^2({\bf y}, \theta_0^{(t)}) & \equiv & \sum_k {\textrm E}\left [ l_{2k}^2 \big{|} {\bf y}, \theta_0^{(t)} \right ] \nonumber \\
\Delta_o^2({\bf y}, \theta_0^{(t)}) & \equiv & \sum_k {\textrm E}\left [ l_{2k+1}^2 \big{|} {\bf y}, \theta_0^{(t)} \right ] \nonumber 
\end{eqnarray}
EM then comes down to iteratively repeating two steps:

\begin{itemize}
\item Expectation \\
Compute 
\begin{displaymath}
P_e, \, P_o, \, \Delta_e^2, \, \Delta_o^2
\end{displaymath}
\item Maximization 
\begin{eqnarray*}
\theta_0^{(t+1)} \, = \, {\textrm arg \,\, max}_{\theta} && \hspace{-.6cm}
\{ -\frac{1}{2} ( \Delta_e^2 \cos^2\theta  + \Delta_o^2\sin^2\theta ) \\
& & \hspace{-.2cm} + P_e \cos \theta  +P_o \sin \theta  \} 
\end{eqnarray*}
\end{itemize}
\vspace{0.5cm}

{\em Kalman} filter \cite{Kalman:1960} provides an efficient way of 
updating conditional expectations, as new
observations become available, in linear-gaussian dynamical models. 
Despite our problem not fitting, nominally, into this category, 
we can still make
use of the algorithm: (\ref{GaussianPrior}) plays the role of an initial
distribution;
as we move along the $1D$ array, with trivial dynamics, the data
observed at the new site becomes available, and the posterior distribution 
$\rho({\bf l} | {\bf y}, {\bf \theta}_0^{(t)})$ is updated accordingly.\\

It may be easily verified that $L(\theta ; \theta_0^{(t)})$ is a quasi-concave function. A binary search suffices to find its global maximum with any desired accuracy.\\

\subsection{Piece-wise constant color}

So far we have assumed that color is constant across the system. Of course, this is an oversimplification, and our next step will be to relax it: we will now demand only that color be piece-wise constant.\\

We recall the way we have recast color demosaicing 
as a blind linear inverse problem:
\begin{eqnarray} \label{FactorAnalysisII}
{\bf y} = \mathcal{D}({\bf \theta}) \, {\bf l}  + {\bf \epsilon} 
\end{eqnarray}
A learning algorithm for the loading matrix ${\mathcal D} = {\textrm diag}(\dots, cos \theta_{2k}, sin \theta_{2k+1}, \dots)$ must now be capable of automatically partitioning the $1D$ array into patches of constant color. Of course, neither the number of patches nor their positions and sizes are known beforehand. To achieve this goal we need a suitable regularizer. Borrowing ideas from robust statistics, we model color as a pairwise Markov random field 
\begin{eqnarray} \label{PieceWise}
\rho({\bf \theta} ; {\bf \beta}) = \frac{1}{{\mathcal Z}({\bf \beta})} \prod_{<i j> \in {\mathcal S}} e^{-\beta_{i j} \,\,  | \sin( \theta_i - \theta_j ) |} 
\end{eqnarray}
${\mathcal S}$ stands for the neighborhood system of the Markov field, and ${\bf \beta} = \{ \beta_{i j}, i,j \in {\textrm Z}\}$ quantify the strength of the link between pairs of neighbors. For the time being we assume that both ${\mathcal S}$ and  ${\bf \beta}$ are known. ${\mathcal Z}({\bf \beta})$ is a normalization constant usually referred to as the 
partition function in the statistical physics literature. To get some insight into (\ref{PieceWise}) notice that 
\begin{eqnarray} \label{TwoWayColorFilter}
 | {\textrm sin}(\theta_i - \theta_j ) | = | {\bf u}_i \times {\bf u}_j|
\end{eqnarray} 
where ${\bf u}_j, j \in {\textrm Z}$ are unitary vectors parallel to ${\bf v}_j$ at each site. The cross-product promotes configurations in which such vectors are aligned, while the laplacian distribution of fluctuations promotes isolated discontinuities \cite{Simoncelli:1996}. \\

Putting together the gaussian model for brightness, (\ref{GaussianPrior}), and (\ref{PieceWise}), we end up with the following joint distribution
\begin{eqnarray} \label{ImageModel}
\rho({\bf y}, {\bf l}, {\bf \theta}) & = & \rho({\bf y} |  {\bf l},  {\bf \theta}) \,\, \rho({\bf l}) \,\, \rho({\bf \theta}) \nonumber 
\end{eqnarray}
where
\begin{eqnarray} \label{DataModelII}
\rho({\bf y} | {\bf l}, {\bf \theta})  & = &{\mathcal N}({\bf y} ; {\mathcal D}({\bf \theta})  {\bf l} , \, \sigma^2 {\bf I})  
 \end{eqnarray}
Pay attention that we are still assuming that color and brightness are marginally independent. In due course we will drop such assumption.\\

To learn color we again resort to expectation-maximization. It now comes down to iteratively optimizing the following function
\begin{eqnarray} \label{PieceWiseFunc}
\sigma^2 L({\bf \theta} ; {\bf \theta}^{(t)}) & = & L_0 -\frac{1}{2} \sum_{j \in {\textrm Z}} h_j^2(\theta_j) \Delta^2_j + \sum_{j \in {\textrm Z}} h_j(\theta_j) P_j \nonumber \\
& & - \sum_{<i, j> \in {\mathcal S}} \sigma^2 \beta_{i j} | {\bf u}_i \, \times \, {\bf u}_j | 
\end{eqnarray}
with
\begin{eqnarray} 
h_j(\theta_j) =  \left \{
\begin{array}{ccc}
\cos\theta_j & , & {\textrm for} \,\, j = 2k\\
\sin\theta_j & , & {\textrm for} \,\, j = 2k+1\\
\end{array}
\right .  \nonumber
\end{eqnarray}
and 
\begin{eqnarray} \label{ConditionalExpect}
P_j \equiv y_{j} {\textrm E}\left [ l_{j} \big{|} {\bf y}, {\bf \theta}^{(t)} \right ] \\
\Delta^2_j \equiv {\textrm E}\left [ l_{j}^2 \big{|} {\bf y}, {\bf \theta}^{(t)} \right ]  \nonumber 
\end{eqnarray}

\begin{figure*}[!ht]
\centering
$\begin{array}{c}
\includegraphics[width=6cm, height=13cm, angle=-90]{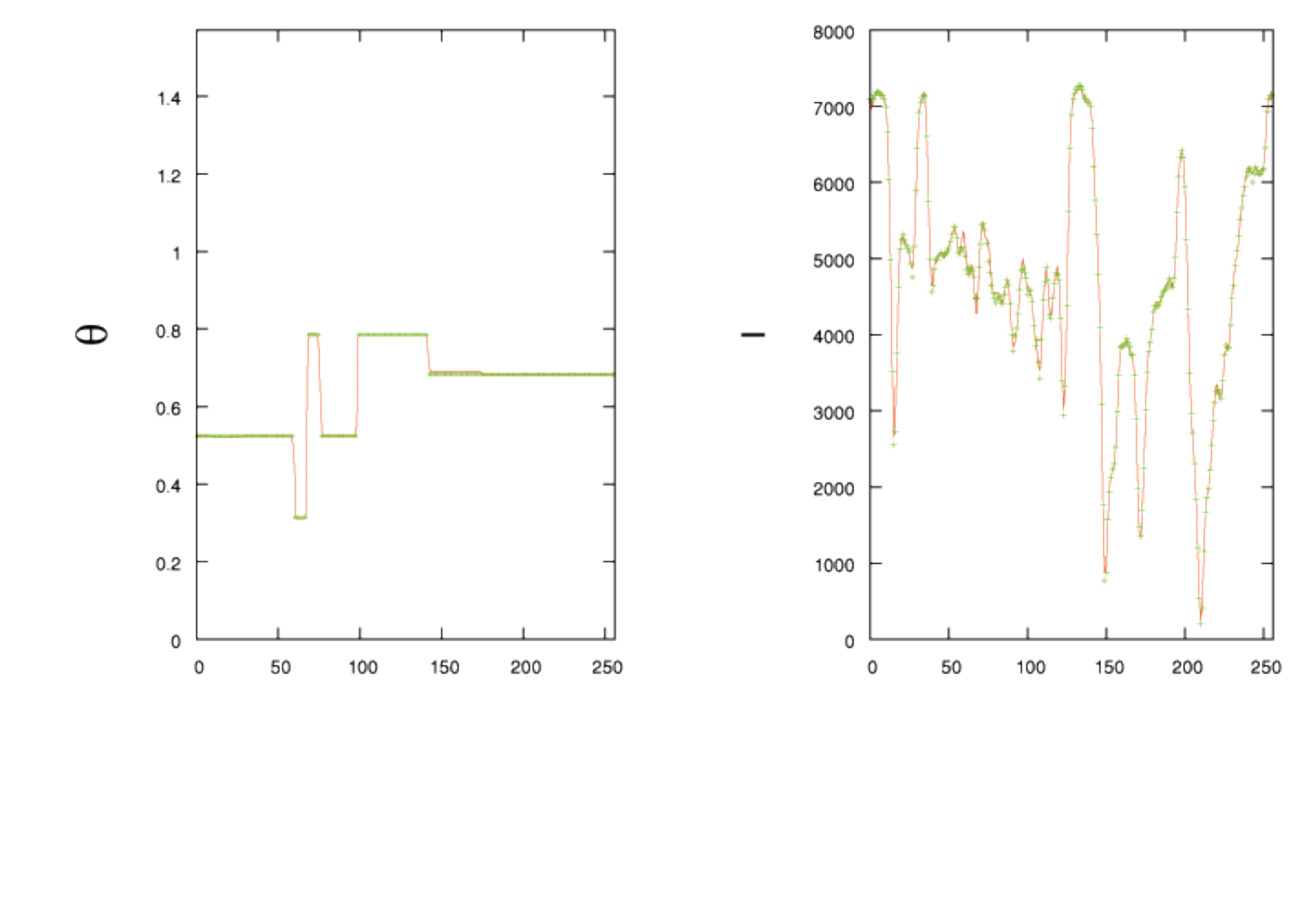} \\
\includegraphics[width=6cm, height=13cm, angle=-90]{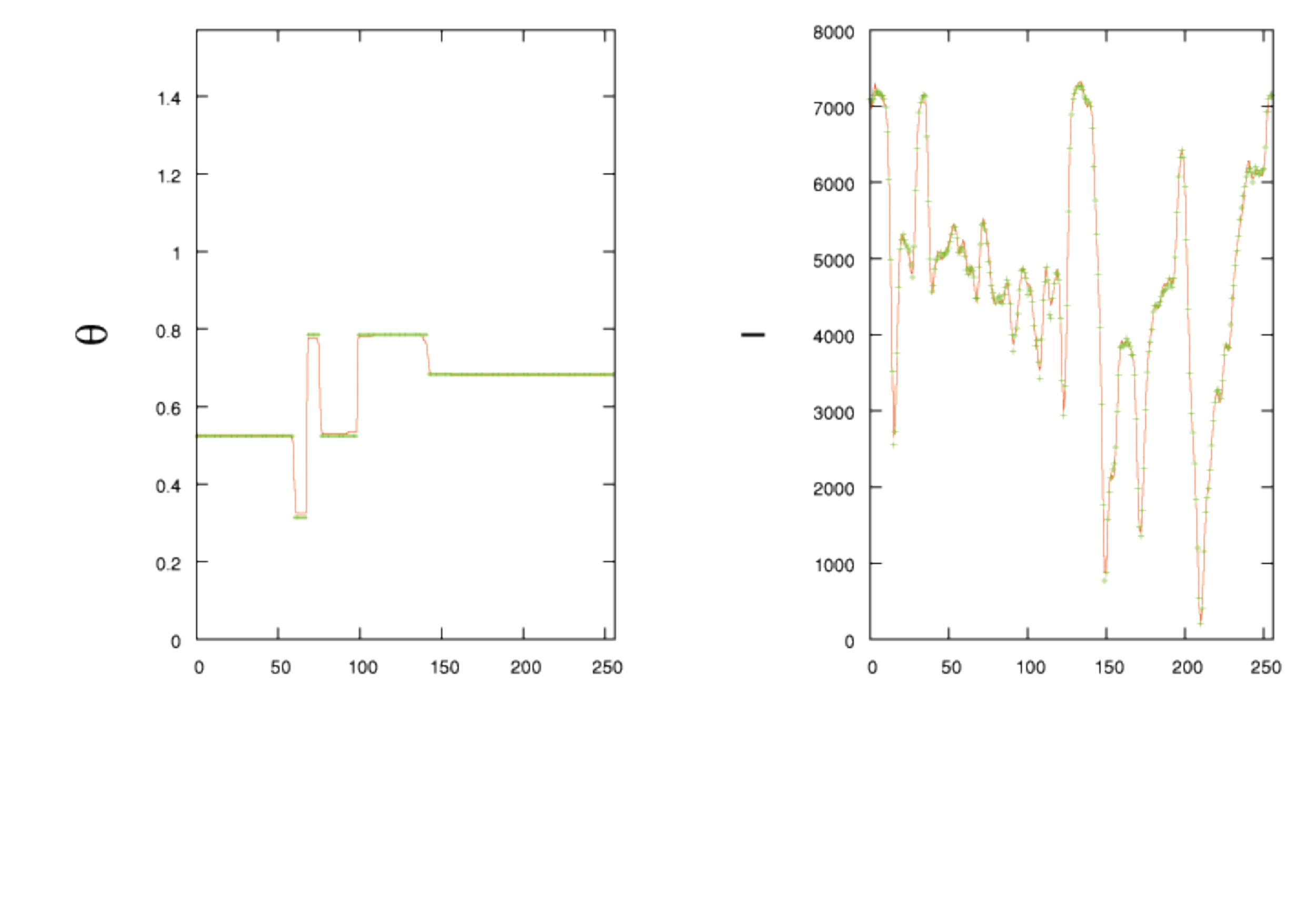} \\
\end{array}$
\caption {Reconstruction of color-brightness in a $1D$ toy problem where color is piecewise constant. The plots on the left compare the reconstructed color (continuous red line) to its original value (green crosses). The plots on the right compare reconstructed brightness (continuous red line) to its original value (green crosses). Two noise levels have been simulated, namely, $\sigma = 30 \, {\textrm (top)}, 50 \, {\textrm (bottom)}$. Both brightness and noise level are measured in arbitrary units.} \label{Color-brightness.B}
\end{figure*}

To validate our method we have carried out some simulations. The study was conducted as follows: we first extracted a linear profile along 256 pixel sites from a textured region of a gray scale image. This profile served as the brightness signal we'll later aim to reconstruct, $\{ l_j, j = 1, 2, \dots, 256\}$. A textured region was selected so that high-frequency structure is present in the profile. Next, we partitioned the linear array into several patches of different lengths, and assigned a constant color to each one. Then cropped the {\em red} and {\em green} projections in alternate sites, and synthetically added white, additive, zero-mean Gaussian noise. Noise levels were chosen to match realistic signal-to-noise ratios for ISOs up to ISO 800. Fig. (\ref{Color-brightness.B}) illustrates the results of the study.\\

For this study we used a uniform regularizer with only nearest neighbors interactions, {\em i.e.} $\beta_{i j} = \beta_0 > 0$ for $|i - j| = 1$, and $\beta_{i j} = 0$, otherwise. The value of $\beta_0$ was handpicked. As the graph of this 
Markov chain contains no loops, a global  optimizer of a discrete version of (\ref{PieceWiseFunc}) can be found using {\em Viterbi's} algorithm \cite{Kschischang:2001}. Regarding the expectation step (\ref{ConditionalExpect}), no changes are needed to cope with
the new color model. As we did in the previous section, we again compute it 
using {\em Kalman} filter.\\

As the plots show, the simple regularizer used in the simulations suffices to learn models of piece-wise constant color.  As it turn out, it's not rich enough to successfully demosaic color images. We tackle this issue next. 
  
%\begin{figure*}[!ht]
%\centering
%$\begin{array}{c}
%\includegraphics[width=6cm, height=13cm, angle=-90]{ToySystemColor/graph_A_30sv.pdf} \\
%\includegraphics[width=6cm, height=13cm, angle=-90]{ToySystemColor/graph_A_45sv.pdf} \\
%\includegraphics[width=6cm, height=13cm, angle=-90]{ToySystemColor/graph_A_60sv.pdf} \\
%\end{array}$
%\caption {Reconstruction of color-brightness in a $1D$ toy problem where color is piecewise constant. The plots on the left compare the reconstructed color (continuous red line) to its original value (dashed green line). The plots on the right compare reconstructed brightness (continuous red line) to its original value (dashed green line). Three noise levels have been simulated, namely, $\sigma = 30 \, {\textrm (top)}, 40 \, {\textrm (middle)}, 50 \, {\textrm (bottom)}$. Both brightness and noise level are measured in arbitrary units.} \label{Color-brightness.C}
%\end{figure*}

\subsection{An adaptive color model}

In the previous section we hypothesized that the neighborhood of each pixel, as well as the strength of the links to its neighbors, ${\{ \beta_{i j}, <i j> \, \in {\mathcal S} \}}$, were known. In this section and the next we show how learned graph structures are plugged into the framework:\\

In the Bayesian spirit we accept uncertainty in the graph structure of the pairwise Markov field by letting ${\bf \beta} = \{ \beta_{i j}, \, i,j \in {\textrm Z} \}$ be themselves random variables\footnote{to be fully general, we should let the MRF include cliques involving more than just two lattice
sites. ${\mathcal S}$ would then designate the set of all allowed cliques,
and ${\bf \beta} = \{ \beta_{C}, \, C \in {\mathcal S} \}$. 
Appropriate potentials for the larger cliques, instead of (\ref{TwoWayColorFilter}), would then have to be designed or learned too}. The joint distribution now factorizes as
\begin{eqnarray} \label{Adaptive}
\rho( {\bf y}, {\bf l},  {\bf \theta} , {\bf \beta} ) = \rho({\bf y} | {\bf l},  {\bf \theta} ) \, \rho({\bf l}) \, \rho( {\bf \theta} | {\bf \beta})  \,  \rho({\bf \beta}) 
\end{eqnarray}
with $\rho({\bf y} | {\bf l},  {\bf \theta} ), \rho({\bf l})$ given respectively by (\ref{DataModelII}), (\ref{GaussianPrior}), and $\rho( {\bf \theta} | {\bf \beta})$ given by (\ref{PieceWise}). Fortunately, we don't need to make any explicit choice for $\rho({\bf \beta})$, neither compute the {\em partition function} ${\mathcal Z}({\bf \beta})$. Let's see why:\\

Expectation-Maximization learns color by repeatedly optimizing a subrogate function that, given (\ref{Adaptive}), (\ref{PieceWise}), looks like 
\begin{eqnarray} \label{FinalSubrogate}
L({\bf \theta} ; {\bf \theta}^{(t)}) &=& L_0+ {\textrm E}\left [ {\textrm log} \, \rho( {\bf y} | {\bf l},  {\bf \theta} ) \big{|} {\bf y}, {\bf \theta}^{(t)} \right ]  \\
& & - \sum_{<i j> \in {\mathcal S}} {\textrm E}\left [  \beta_{i j} \big{|} {\bf y}, {\bf \theta}^{(t)} \right ]  |\sin(\theta_i - \theta_j )|  \nonumber 
\end{eqnarray}
where the only change with respect to (\ref{PieceWiseFunc}) is that we have made the replacement
\begin{eqnarray}
\beta_{i j} \longrightarrow  {\textrm E}\left [  \beta_{i j} \big{|} {\bf y}, {\bf \theta}^{(t)} \right ]  \nonumber
\end{eqnarray}
Under model (\ref{Adaptive}), this conditional expectation happens to be independent of the data, ${\bf y}$ 
\begin{eqnarray}
{\textrm E}\left [  \beta_{i j} \big{|} {\bf y}, {\bf \theta}^{(t)}  \right ]  & = & \int d{\bf l} d{\bf \beta} \, \beta_{i j} \, \rho({\bf l}, {\bf \beta} \big{|} {\bf y}, {\bf \theta}^{(t)})  \nonumber 
\end{eqnarray}
\begin{eqnarray}
& = & \frac{\int d{\bf l} d{\bf \beta} \, \beta_{i j} \, \rho({\bf y}, {\bf l}, {\bf \theta}^{(t)}, {\bf \beta})}{\int d{\bf l} d{\bf \beta}  \, \rho({\bf y}, {\bf l}, {\bf \theta}^{(t)}, {\bf \beta})} \nonumber \\ 
& = & \frac{\int d{\bf l} \,\, \rho({\bf y} \big{|} {\bf l}, {\bf \theta}^{(t)}) \, \rho({\bf l}) \, \int d{\bf \beta} \, \beta_{i j} \, \rho({\bf \theta}^{(t)} ,  {\bf \beta} ) }{ \int d{\bf l} \,\, \rho({\bf y} \big{|} {\bf l}, {\bf \theta}^{(t)}) \, \rho({\bf l}) \, \int d{\bf \beta} \, \rho({\bf \theta}^{(t)},  {\bf \beta} ) } \nonumber \\
& = & \int d{\bf \beta} \, \beta_{i j} \, \rho( {\bf \beta}  \big{|} {\bf \theta}^{(t)})  \nonumber \\
& = & {\textrm E}\left [  \beta_{i j} \big{|} {\bf \theta}^{(t)} \right ]  \nonumber
\end{eqnarray}
Two important points need to be stressed here:
\begin{itemize}
\item [] a) we just need the expected value of the posterior distribution $\rho({\bf \beta} \big{|} \, {\bf \theta})$, and not the distribution function itself
\item [] b) being independent of the data, a regression model for this 
conditional expectation may be learned in advance
\end{itemize}

\subsection{Color-Brightness cross-dependencies}

So far (see (\ref{Adaptive})),  we have assumed that color and brightness are marginally independent. This might seem a reasonable assumption given that brightness often changes at a fast pace in regions where color remains constant.  Actually, it is not: steep changes in color are always accompanied by steep changes in brightness, and failure to have these boundaries registered results in highly visible artifacts. \\

We may still capture color-brightness dependencies within a workable learning scheme if we just assume that they are fully encompassed in the graph structure of the image model, {\em i.e.}
\begin{eqnarray} \label{HiddenMarkov}
\rho({\bf l}, {\bf \theta} \big{|} {\bf \beta} ) =  \rho({\bf l} \big{|} {\bf \beta} ) \, \rho({\bf \theta} \big{|} {\bf \beta} )
\nonumber
\end{eqnarray}
The joint distribution then factorizes as
\begin{eqnarray} \label{Cross}
\rho( {\bf y}, {\bf l},  {\bf \theta} , {\bf \beta} ) & = & \rho({\bf y} | {\bf l},  {\bf \theta} ) \, \rho({\bf l} \big{|} {\bf \beta} ) \\
& & \times \,\, \rho( {\bf \theta} | {\bf \beta}) \, \rho({\bf \beta}) \nonumber
\end{eqnarray}

What is left is to derive a learning algorithm for this model. This is done in
the Appendix. Here we just summarize our findings: the entanglement between color and brightness poses a challenge. Fortunately, within a Laplace approximation, (\ref{PieceWiseFunc}) remains valid as long as we replace $\{ \beta_{i j} \}$ with a regression function that now depends on both color and brightness
\begin{eqnarray} \label{Regression}
\beta_{i j} \longrightarrow \hat{\beta}_{i j} ({\bf l}^{(t)}, {\bf \theta}^{(t)})  = {\textrm E}\left [  \beta_{i j} \big{|} {\bf l}^{(t)}, {\bf \theta}^{(t)} \right ] 
\end{eqnarray}
where $ {\bf l}^{(t)} = {\textrm E}^{*} \left [  {\bf l} \big{|} {\bf y}, {\bf \theta}^{(t)} \right ]$; ${\textrm E}^{*}$ stands for the conditional expectation with respect to the gaussian process $\rho({\bf y} | {\bf l},  {\bf \theta} ) \, \rho({\bf l})$.\\

The regression function (\ref{Regression}) enriches the one we introduced in the previous section. Basically, the claim we are making here is that, 
under model (\ref{Cross}), the size and shape
of the region around
a pixel site, say $i$, where color is expected to remain constant, 
${\mathcal S}_i \equiv \{j \in {\textrm Z} : \beta_{i j} > 0 \}$, can be predicted from the color of the surroundings and the intensity of the illumination
there.
Such predictive rule is to be learned in advance in a
supervised manner.\\ 

This completes our learning-inference framework for color demosaicing. The following pseudo-code summarizes the scheme 

\begin{itemize}
\item [0.]  Learn off-line the regression function $\hat{\beta}_{i j} ( {\bf l}, {\bf \theta} ) $
\end{itemize}

To demosaic an image, repeat iteratively the following steps
\begin{itemize}
\item [1.] Expectation: given the current estimate of color, infer brightness  
\begin{eqnarray}
{\bf l}^{(t)} = {\textrm E}^{*} \left [  {\bf l} \big{|} {\bf y}, {\bf \theta}^{(t)} \right ] \nonumber
\end{eqnarray}
as well as the conditional variance of each of its components (as required in (\ref{PieceWiseFunc})). 
\item [2.] Evaluate $\hat{\beta}_{i j} ( {\bf l}^{(t)}, {\bf \theta}^{(t)} )$ at the current estimates of color and brightness. Plug it into (\ref{PieceWiseFunc}). 
\item [3.] Maximization: update the estimate of color by optimizing (\ref{PieceWiseFunc}) 
\end{itemize}

%\IEEEpubidadjcol

%\subsubsection{Subsubsection Heading Here}
%Subsubsection text here.

\section{Demosaicing color images}

In the previous section we presented our framework for color demosaicing in the context of the {\em toy problem}       depicted in Fig.\ref{ToyProblem}. Demosaicing actual images requires some adjustments. In this section we review them:
\begin{itemize}
\item [a.] Whereas our {\em toy} problem comprised just two color separations, and
a single angle, $\theta$,  was enough to specify their ratio at each site,
color images comprise three
separations. Color will now be represented by a $3D$ unitary vector taking values on the
positive octant of the sphere, or, equivalently, by two angles, $\theta, \phi \in [0, \pi/2]$. 
Entries in the diagonal loading matrix ${\mathcal D}({\bf \theta, \phi}) = {\textrm diag}(\dots, h_j(\theta_j, \phi_j), \dots)$ need to be replaced with
\begin{eqnarray} \label{CFA}
h_j(\theta_j, \phi_j) =  \left \{
\begin{array}{ccc}
\cos\theta_j & , & {\textrm for} \,\, j \in {\mathcal G}  \\
 \sin\theta_j \, \cos \phi_j & , & {\textrm for} \,\, j \in {\mathcal R} \\
 \sin\theta_j \, \sin \phi_j & , & {\textrm for} \,\, j \in {\mathcal B}
\end{array}
\right .  \nonumber
\end{eqnarray}
where ${\mathcal G}, {\mathcal R}, {\mathcal B} \subset {\textrm Z}^2$ stand for the sublattices of the color filter array where {\em green}, {\em red} and {\em blue} are, respectively, measured; and (\ref{PieceWiseFunc}) is replaced with
\begin{eqnarray} \label{L2D}
\sigma^2 L({\bf \theta}, {\bf \phi} ; {\bf \theta}^{(t)}, {\bf \phi}^{(t)} ) & = & -\frac{1}{2} \sum_{j \in {\textrm Z}^2} h_j^2(\theta_j, \phi_j) \Delta^2_j 
+ \sum_{j \in {\textrm Z}^2} h_j(\theta_j, \phi_j) P_j \nonumber \\
& & - \sum_{<i, j> \in {\mathcal S}} \sigma^2 \, \hat{\beta}_{i j}({\bf l}^{(t)}, {\bf \theta}^{(t)}, {\bf \phi}^{(t)}) \, | {\bf u}_i \, \times \, {\bf u}_j | 
\end{eqnarray}
with $P_j = y_{j} {\textrm E}\left [ l_{j} \big{|} {\bf y}, {\bf \theta}^{(t)}, 
{\bf \phi}^{(t)} \right ]$, $\Delta^2_j = {\textrm E}\left [ l_{j}^2 \big{|} {\bf y}, {\bf \theta}^{(t)}, {\bf \phi}^{(t)} \right ]$, and ${\bf u} = (\cos \theta, \sin \theta \cos \phi, \sin \theta \sin \phi )$.\\

\item [b.] Pixel sites in our {\em toy} problem were arranged in a $1D$ chain. Images, on
the other hand, are $2D$. On the surface this seems to have required a very minor adjustment (compare (\ref{L2D}) to (\ref{PieceWiseFunc})).
However, the larger 
system size is going to force upon us some changes at the implementation level. We review
them next: 

\begin{itemize}
\item [] {\bf Expectation Step:} In our simulations of the previous section,
we computed (\ref{ConditionalExpect}) by means of {\em Kalman} filter.
The memory 
requirements
of this algorithm, however, grow quadratically with the number of lattice sites,
and become untenable for our ultimate purposes. A more efficient 
alternative
is provided by low-memory quasi-Newton optimizers \cite{Nocedal:1995}. We
exploit the fact that the average and the mode of a gaussian distribution
coincide, and that its covariance matrix happens to be the inverse
of the hessian of its logarithm. Low-memory quasi-Newton optimizers provide
efficient estimates for both of them. 

\item [] {\bf Maximization Step:} In our simulations of the previous section,
we resorted to {\em Viterbi}'s algorithm to find,
with any desired accuracy, the global optimum of (\ref{PieceWiseFunc}).
This was possible thanks to our readiness there
to model color as a MRF with no loops in its graph (remember, we chose $\beta_{i j} = 0$ for $|i - j| \neq 1$). 
When demosaicing color images we cannot afford this simplification, and must deal with MRF whose graphs do contain loops.
For them, belief propagation algorithms, like {\em Viterbi}'s, cannot guarantee convergence to a
global optima any longer. This is an important distinction, though only at a formal level;
in practice, for the problem at hand and for many others \cite{Gallager:1963}, it happens to be of little relevance. There is, however, another
obstacle that do prevent us from using loopy belief propagation, namely,
the degrees of freedom we aim to estimate, $\theta$ and $\phi$, take values in the continuum, instead
of a discrete set. In the previous section, we bypassed this difficulty
by just discretizing the interval $[0, \pi/2]$ finely enough. With
color images, 
this is simply not viable.
Sophisticated extensions, like non-parametric belief propagation \cite{Sudderth:2003},
 could still be considered. Fortunately, we don't need them, as
the simplest of the approaches suffices to do the job:
in the spirit of generalized EM \cite{Wu:1983}, we sequentially solve univariate 
optimization problems on each of the degrees of freedom involved.\\

\end{itemize}

\item [c.] To initialize our whole procedure we have
chosen, for its simplicity, Laroche's {\em gradient-based} interpolation 
\cite{Laroche:1994}. More sophisticated interpolations,
like those described in \cite{Hamilton:1997} and {Chang:1999}, may certainly 
yield
improved results. Such considerations are, however, beyond the scope
of our current interest.\\

\end{itemize}

\section{Results}

Our primary goal so far has been the development of a  
framework for color demosaicing
that may enable the solution of this problem in a systematic manner.
We presented such framework in the preceeding sections.
The main conclusion to be drawn from the ideas
exposed there is the following:
color demosaicing is possible inasmuch as a regressor $\hat{\bf \beta}({\bf l}, {\bf \theta}, {\bf \phi})$ can be
learned, {\em i.e} as long as the size and shape of 
the local region where
color is expected to remain constant can be predicted from the color of
the surroundings and the conditions of illumination there. 
Our scheme then provides a principled prescription for the 
integration of that
knowledge into an adaptive filter \footnote{implicitly defined as the outcome,
at each iteration, of our expectation-maximization algorithm}, similar in
spirit to those
proposed by Lukac {\em et al} \cite{Lukac:2005}.\\

What is still missing for our framework to be complete is a recipe 
for learning $\hat{\bf \beta}({\bf l}, {\bf \theta}, {\bf \phi})$. 
The need for such procedure is made more acute by the realization that
no single regression model
might be suitable for all sensors and kinds of illuminants, 
irrespective of the spectral response of the former and the spectral content
of the latter.
Ideally then, whenever any of these two changes,
$\hat{\bf \beta}({\bf l}, {\bf \theta}, {\bf \phi})$ 
should be learned anew, 
so that it trully captures the joint statistics
of the three color separations that are actually 
presented to the demosaicing algorithm. Efforts on this front are underway 
and will be reported elsewhere.\\ 

This said, it is evident that only empirical evidence can bring  
true plausibility to our approach. To provide it, we 
have crafted a regression model,
with the help of feedback from human experts, and used it in a 
comparative study of performance with
a well-known, state-of-the-art demosaicing algorithm. 
For the sake of completeness,
we review the most remarkable features of this model of ours.
\begin{eqnarray} \label{HandCraft}
\beta_{i, j}( {\bf l}, {\bf \theta}, {\bf \phi}) = \beta_0 \, e^{-\frac{1}{2} |{\bf r}_i - {\bf r}_j|^2/d^2({\bf l}, {\bf \theta}, {\bf \phi})} \, (1 + \alpha \, e^{-|l_i - l_j|/R})
\end{eqnarray}
$\alpha$ and $R$ are constants; $|{\bf r}_i - {\bf r}_j|$ is the euclidean distance between lattice sites
$i$ and $j$; 
$d({\bf l}, {\bf \theta}, {\bf \phi})$ sets the scale of the local region 
around each pixel 
where no color modulations are expected to occur. Its complex 
functional dependece on its arguments, $\{{\bf l}, {\bf \theta}, {\bf \phi}\}$, we have encoded in 
a regression tree \cite{Breiman:1984} that we learned as follows:\\
we first assembled a collection of patches of uniform color, and demosaiced 
them using various different values for the scale $d$. 
With the help of human experts, one of those values was chosen as optimal
for the patch, and recorded.
In addition to this scale,
we also recorded, for selected pixels in the patch and for windows 
of increasing size around them, the maximum and minimum values
of the {\em green} channel, and the 
{\em red-to-green} and {\em blue-to-green} differences. The latter 
are meant to serve as prediction attributes for $d$. 
Our regression tree was then learned from all this data.
One issue must be stressed: the prediction attributes fed to the
learning algorithm were collected from demosaiced patches. At no point 
whatsoever were the original color pictures, even if available,
shown to it. Some 
remarkable aspects
of our learned predictor are worth mentioning
\begin{itemize}
\item [1.] {\em Non-locality}: Windows of up to $20\times20$ pixels are needed
to accurately predict $d$, even though $d$ itself takes on much smaller
values,   
typically in the interval $\sim (0.25, 3)$ pixels. 
\item [2.] {\em Color Saturation}: On backgrounds of saturated colors, 
$d$ takes its smallest values, $\sim 0.25$ pixels; on non-saturated backgrounds, (where
the three color separations take almost equal values), its largest, $\sim 3.$ 
pixels. In other words, 
color
modulations are effectively quenched on backgrounds of the latter kind. 
This, in turn, 
should imply that 
the photosensor array, and possibly our visual system too, is most 
sensitive to brightness
modulations that occur on top of backgrounds of such colors\footnote{for that matter, this might be the reason why this paper is written on a white sheet, instead of a vivid red one}. 
\item [3.] {\em Brightness Saturation} On very bright regions,
$d$ takes large values. This is consistent with the behaviour of 
devices, like monitors, that saturate as brightness surpass certain
threshold. In such conditions thay are rendered unable to reproduce 
color modulations.
\item [4.] {\em Color/Brightness Edges}. Sharp changes in color are almost
systematically accompanied by sharp changes in brightness, or, equivalently,
an absence of changes in brightness is likely to imply that color remains constant. This
behaviour is captured by the second factor in (\ref{HandCraft}). It is this
term that confers shape to the local region around a pixel where color constancy holds.\\
\end{itemize}

Our comparative study of performance follows on the footsteps of that 
carried out in \cite{Li:2008}: 
20 pictures were selected from the {\em PhotoCD PCD0992} set of high
quality color images \cite{UC-SIPI}. These are scanned analogical pictures,
for which the true values of the three color separations are known.
Data was cropped
from them as if seen by a Bayer color filter array, and fed into our
algorithm for demosaicing. To serve as references the same data was also fed
to Li's successive approximation \cite{Li:2005}, SA.
As in \cite{Li:2008}, two quantitative measures of
performance are provided: PSNR and SCIELAB \cite{SCIELAB}. 
Neither of these two metrics, however, can 
trully measure how perceptually pleasant
a demosaiced image is to a human viewer \cite{Longere:2002}.
For this reason, as
it is customary, we report as well subjective assesments of performance.\\

\begin{figure*}[!ht]
\centering
$\begin{array}{cccc}
\includegraphics[width=3.7cm, height=3.7cm, angle=0]{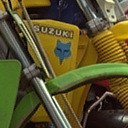} \hspace{0.25cm}
\includegraphics[width=3.7cm, height=3.7cm, angle=0]{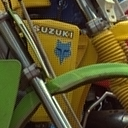} \hspace{1cm}
\includegraphics[width=3.7cm, height=3.7cm, angle=0]{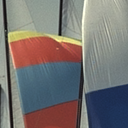} \hspace{0.25cm}
\includegraphics[width=3.7cm, height=3.7cm, angle=0]{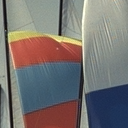} \\
\includegraphics[width=3.7cm, height=3.7cm, angle=0]{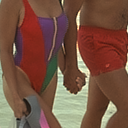} \hspace{0.25cm}
\includegraphics[width=3.7cm, height=3.7cm, angle=0]{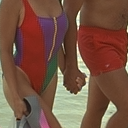} \hspace{1cm}
\includegraphics[width=3.7cm, height=3.7cm, angle=0]{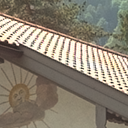} \hspace{0.25cm}
\includegraphics[width=3.7cm, height=3.7cm, angle=0]{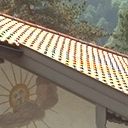} \\
\includegraphics[width=3.7cm, height=3.7cm, angle=0]{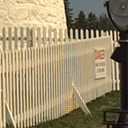} \hspace{0.25cm}
\includegraphics[width=3.7cm, height=3.7cm, angle=0]{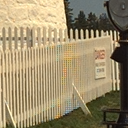} \hspace{1cm}
\includegraphics[width=3.7cm, height=3.7cm, angle=0]{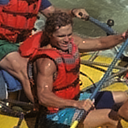} \hspace{0.25cm}
\includegraphics[width=3.7cm, height=3.7cm, angle=0]{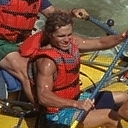} \\
\end{array}$
\caption {Illustrative examples of pictures demosaiced using the algorithm described in this paper ({\em left}). For comparisson, pictures demosaiced 
from the same data by succesive approximation (SA) are also shown ({\em right}).
} \label{Pictures}
\end{figure*}

Tables \ref{SCIELAB_8bits} and \ref{PSNR_8bits}, and 
Figure \ref{Pictures} illustrate the results of our study, whose 
conclusions we
summarize as follows: 
SA consistenly scores higher than our algorithm both in SCIELAB and PSNR.
On the other hand, our demosaiced images are free from 
the very evident color artifacts that plague SA reconstructions.\\  

\begin{figure*}[!ht]
\centering
$\begin{array}{cc}
\includegraphics[width=8cm, height=8cm, angle=0]{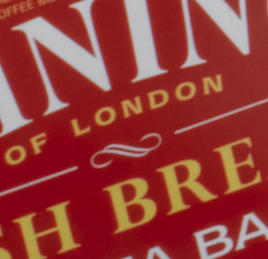} \hspace{0.5cm}
\includegraphics[width=8cm, height=8cm, angle=0]{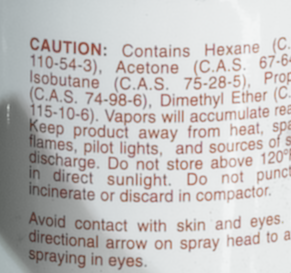}\\
\end{array}$
\caption {Illustrative examples of pictures demosaiced from digitally captured
data by the algorithm described in this paper. To avoid aliasing artifacts
when viewing the pictures in a monitor, please enlarge the images} 
\label{Pictures2}
\end{figure*}

For a better appreciation of the possibilities of the ideas described in
this paper, we provide in Fig.\ref{Pictures2} examples of images demosaiced
with our algorithm from digitally captured data. To deal with this kind
of data, a new regression tree for $d({\bf l}, {\bf \theta}, {\bf \phi})$ 
needed to be learned.\\

\begin{table*}[!ht]
\caption{$SCIELAB$ \textit{for images demosaiced from data known with a precision of 8 bits.}}
\begin{center}
\begin{tabular}{|c|c|c|c|c|c|c|c|c|c|c|}
\hline
kodim & 01 & 02 & 03 & 04 & 05 & 06 & 07 & 08 & 09 & 10 \\
\hline
Ours & 1.11 & 1.01 & 0.64 & 0.64 & 1.31 & 0.94 & 0.49 & 1.21 & 0.61 & 0.57 \\
\hline
SA & 0.99 & 0.73 & 0.60 & 0.64 & 1.32 & 0.89 & 0.48 & 1.23 & 0.63 & 0.56 \\
\hline
\end{tabular}

\vspace{.25cm}

\begin{tabular}{|c|c|c|c|c|c|c|c|c|c|c|}
\hline
kodim & 11 & 12 & 15 & 16 & 19 & 20 & 21 & 22 & 23 & 24 \\
\hline
Ours & 0.98 & 0.47 & 0.78 & 0.56 & 0.98 & 0.63 & 1.04 & 0.98 & 0.47 & 1.02 \\
\hline
SA & 0.86 & 0.49 & 0.77 & 0.54 & 0.90 & 0.62 & 0.92 & 1.00 & 0.52 & 1.03\\
\hline
\end{tabular}

\end{center}
\label{SCIELAB_8bits}
\end{table*}

\begin{table*}[!ht]
\caption{$PSNR$ \textit{for images demosaiced from data known with a precision of 8 bits.}}
\begin{center}
\begin{tabular}{|c|c|c|c|c|c|c|c|c|c|c|}
\hline
kodim & 01 & 02 & 03 & 04 & 05 & 06 & 07 & 08 & 09 & 10\\
\hline
Ours & 64.57 & 64.52 & 64.93 & 66.00 & 64.16 & 64.65 & 66.20 & 64.67 & 66.26 & 66.56 \\
\hline
SA & 66.96 & 66.61 & 67.50 & 67.67 & 65.70 & 66.66 & 68.57 & 65.72 & 68.18 & 68.12\\
\hline
\end{tabular}

\vspace{.25cm}

\begin{tabular}{|c|c|c|c|c|c|c|c|c|c|c|}
\hline
kodim & 11 & 12 & 15 & 16 & 19 & 20 & 21 & 22 & 23 & 24\\
\hline
Ours & 64.80 & 67.21 & 64.76 & 65.25 & 65.99 & 65.88 & 64.80 & 65.07 & 66.23 & 64.35\\
\hline
SA & 67.08 & 68.57 & 66.24 & 67.99 & 67.15 & 67.07 & 66.97 & 66.36 & 67.48 & 64.84\\
\hline
\end{tabular}

\end{center}
\label{PSNR_8bits}
\end{table*}

\section{Conclusions}

We have presented a statistical learning framework for color demosaicing. 
The problem was recast as a blind linear inverse one: color parameterizes
the unknown kernel; brightness takes on the role of a latent variable.
An expectation-maximization algorithm was designed to learn the former in
an unsupervised manner. The latter is then inferred from the data.\\

Our learning algorithm is based on a pairwise Markov random field representation of color similarities. Non-local color correlations, as well as those 
between color and brightness, are absorbed into the graph structure of the Markov field. Letting this structure change as a function of the background's color and illumination makes the whole procedure adaptive.\\

The aforementioned adaptive behaviour needs to be learned in advance. 
Even more, we claim demosaicing is possible as long as such adaptive
bahaviour can be learned.
Our framework then readily accepts such {\em prior} knowledge in a {\em plug-and-play} fashion, through the regression function ${\bf \beta}({\bf l}, {\bf \theta}, {\bf \phi})$. \\

We have proposed efficient algorithms to carry out all but one of the subtasks
in our expectation-maximization scheme. The missing piece is a procedure
for learning, in an systematic manner,
 ${\bf \beta}({\bf l}, {\bf \theta}, {\bf \phi})$. Progress on that track will be reported elsewhere.\\

% or


\appendix[]
%\section{}

In this appendix we present a full derivation of the subrogate function $L({\bf \theta} ; {\bf \theta}^{(t)})$ associated to model (\ref{Cross}). ${\bf \theta}^{(t)}$ stands for the 
current estimate of color after {\em t} iterations
\begin{eqnarray}
L({\bf \theta} ; {\bf \theta}^{(t)}) & = & 
{\textrm E}\left [  \log \rho( {\bf y}, {\bf l}, {\bf \theta}, {\bf \beta} ) \big{|} {\bf y}, {\bf \theta}^{(t)} \right ] \nonumber \\
& = & L_0 + {\textrm E}\left [  \log \rho( {\bf y} | {\bf l}, {\bf \theta}) \big{|} {\bf y}, {\bf \theta}^{(t)} \right ] \nonumber \\
&   &  +  {\textrm E}\left [  \log \rho( {\bf \theta} \big{|} {\bf \beta}) \big{|} {\bf y}, {\bf \theta}^{(t)} \right ] \nonumber 
\end{eqnarray}
We start with the second contribution to $L({\bf \theta} ; {\bf \theta}^{(t)})$ . Given (\ref{PieceWise}), it reads
\begin{eqnarray*}
{\textrm E}\left [  \log \rho( {\bf \theta} \big{|} {\bf \beta}) \big{|} {\bf y}, {\bf \theta}^{(t)} \right ]  & = & \hspace{3cm} 
\end{eqnarray*}
\begin{eqnarray*}
& = &  - \sum_{<i j> \in {\mathcal S}} {\textrm E}\left [  \beta_{i j} \big{|} {\bf y}, {\bf \theta}^{(t)} \right ]  |\sin(\theta_i - \theta_j )|  \nonumber 
\end{eqnarray*}
In order to bring ${\textrm E}\left [  \beta_{i j} \big{|} {\bf y}, {\bf \theta}^{(t)} \right ]$ into a desirable expression, we make the following approximation: as a function of ${\bf l}$, $\rho({\bf y} \big{|} {\bf l}, {\bf \theta}^{(t)}) \, \rho({\bf l}) $ is a gaussian centered at
\begin{eqnarray}
{\bf l}^{(t)} = {\textrm E}^{*} \left [ l_j  \big{|} {\bf y}, {\bf \theta}^{(t)} \right ] = \frac{\int d{\bf l} \,\, {\bf l} \, \rho({\bf y} \big{|} {\bf l}, {\bf \theta}^{(t)}) \, \rho({\bf l}) }{\int d{\bf l} \,\, \rho({\bf y} \big{|} {\bf l}, {\bf \theta}^{(t)}) \, \rho({\bf l}) }  \nonumber 
\end{eqnarray} 
where ${\textrm E}^{*} \left [ l_j  \big{|} {\bf y}, {\bf \theta}^{(t)} \right ] $ stands for a conditional expectation with respect to the gaussian process $\rho({\bf y} \big{|} {\bf l}, {\bf \theta}^{(t)}) \, \rho({\bf l})$. 
Under the assumption that $\rho({\bf \beta} \big{|} {\bf l})$ varies slowly as a function of ${\bf l}$ (relative to the spread of $\rho({\bf y} \big{|} {\bf l}, {\bf \theta}^{(t)}) \, \rho({\bf l})$), we approximate ${\textrm E}\left [  \beta_{i j} \big{|} {\bf y}, {\bf \theta}^{(t)} \right ]$ as follows
\begin{eqnarray*}
{\textrm E}\left [ \beta_{i j}  \big{|} {\bf y}, {\bf \theta}^{(t)} \right ] & = & \int d{\bf \beta} \, \beta_{i j} \, \rho({\bf \beta} \, \big{|}  {\bf y}, {\bf \theta}^{(t)}) 
\end{eqnarray*}
\begin{eqnarray*}
& = & \frac{\int d{\bf l} d{\bf \beta} \,\, \beta_{i j} \, \rho({\bf y}, {\bf l}, {\bf \theta}^{(t)}, {\bf \beta})}{\int d{\bf l} d{\bf \beta} \,\, \rho({\bf y}, {\bf l}, {\bf \theta}^{(t)}, {\bf \beta})}  \nonumber \\
& = & \frac{\int d{\bf l} d{\bf \beta} \,\, \beta_{i j} \, \rho({\bf y} \big{|} {\bf l}, {\bf \theta}^{(t)}) \, \rho({\bf \theta}^{(t)} \big{|} {\bf \beta}) \, \rho( {\bf \beta} \big{|} {\bf l} ) \, \rho({\bf l}) }{\int d{\bf l} d{\bf \beta} \, \rho({\bf y} \big{|} {\bf l}, {\bf \theta}^{(t)}) \, \rho({\bf \theta}^{(t)} \big{|} {\bf \beta}) \, \rho( {\bf \beta} \big{|} {\bf l} ) \, \rho({\bf l})}  \nonumber \\
& \approx & \frac{\int d{\bf l} d{\bf \beta} \,\, \beta_{i j} \, \rho({\bf y} \big{|} {\bf l}, {\bf \theta}^{(t)}) \, \rho({\bf \theta}^{(t)} \big{|} {\bf \beta}) \, \rho( {\bf \beta} \big{|} {\bf l}^{(t)} ) \, \rho({\bf l}) }{\int d{\bf l} d{\bf \beta} \, \rho({\bf y} \big{|} {\bf l}, {\bf \theta}^{(t)}) \, \rho({\bf \theta}^{(t)} \big{|} {\bf \beta}) \, \rho( {\bf \beta} \big{|} {\bf l}^{(t)} ) \, \rho({\bf l})} \nonumber 
\end{eqnarray*}
and, invoking (\ref{HiddenMarkov}), we complete our derivation
\begin{eqnarray*}
& = & \frac{\int d{\bf \beta} \, \beta_{i j} \,\,  \rho({\bf \theta}^{(t)} \big{|} {\bf \beta}) \, \rho( {\bf \beta} \big{|} {\bf l}^{(t)} ) \, \, \int d{\bf l} \, \rho({\bf y} \big{|} {\bf l}, {\bf \theta}^{(t)}) \, \rho({\bf l}) }{\int d{\bf \beta} \,\, \rho({\bf \theta}^{(t)} \big{|} {\bf \beta}) \, \rho( {\bf \beta} \big{|} {\bf l}^{(t)} ) \, \, \int d{\bf l} \, \rho({\bf y} \big{|} {\bf l}, {\bf \theta}^{(t)}) \, \rho({\bf l}) } \nonumber \\
& = & \frac{\int d{\bf \beta} \, \beta_{i j} \,\,  \rho({\bf \theta}^{(t)} \big{|} {\bf \beta}) \, \rho( {\bf l}^{(t)} \big{|} {\bf \beta}) \, \rho({\bf \beta}) }{\int d{\bf \beta} \, \rho({\bf \theta}^{(t)} \big{|} {\bf \beta}) \, \rho( {\bf l}^{(t)} \big{|} {\bf \beta}) \, \rho({\bf \beta}) } \nonumber \\
& = & \frac{\int d{\bf \beta} \, \beta_{i j} \,\,  \rho({\bf l}^{(t)}, {\bf \theta}^{(t)} \big{|} {\bf \beta}) \, \rho({\bf \beta}) }{\int d{\bf \beta} \,  \rho( {\bf l}^{(t)}, {\bf \theta}^{(t)} \big{|} {\bf \beta}) \, \rho({\bf \beta}) } \nonumber \\
& = & \int d{\bf \beta} \, \beta_{i j} \,\,  \rho({\bf \beta} \big{|} {\bf l}^{(t)}, {\bf \theta}^{(t)} ) \nonumber \\
& = & {\textrm E}\left [ \beta_{i j}  \big{|} {\bf l}^{(t)}, {\bf \theta}^{(t)} \right ]   \nonumber 
\end{eqnarray*}

The second of the two contributions reads
\begin{eqnarray*}
{\textrm E}\left [  \log \rho( {\bf y} | {\bf l}, {\bf \theta} ) \big{|} {\bf y}, {\bf \theta}^{(t)} \right ]  & = & 
\end{eqnarray*}
\begin{eqnarray*}
& = & {\textrm E}\left [ \log {\mathcal N}({\bf y}; {\mathcal D}({\bf \theta}) {\bf l}, \sigma^2 {\bf I} ) \big{|} {\bf y}, {\bf \theta}^{(t)} \right ]  \nonumber \\
& = & E_0 - \frac{1}{2 \sigma^2}  \sum_j {\textrm E}\left [ (y_j - h_j(\theta_j) l_j )^2  \big{|} {\bf y}, {\bf \theta}^{(t)} \right ] \nonumber \\
& = & E'_0 - \frac{1}{2 \sigma^2}  \sum_j h_j(\theta_j)^2 {\textrm E}\left [ l_j^2  \big{|} {\bf y}, {\bf \theta}^{(t)} \right ] \nonumber \\
& &    + \frac{1}{\sigma^2} \sum_j h_j(\theta_j) y_j {\textrm E}\left [ l_j  \big{|} {\bf y}, {\bf \theta}^{(t)} \right ] \nonumber
\end{eqnarray*}
with
\begin{eqnarray} \label{CFA0}
h_j(\theta_j) =  \left \{ 
\begin{array}{ccc}
\cos \theta_j & , & {\textrm for} \,\, j = 2k  \\
 \sin \theta_j & , & {\textrm for} \,\, j = 2k+1 
\end{array}
\right .  \nonumber 
\end{eqnarray}
For linear-gaussian processes, ${\textrm E}\left [ l_j , \, l_j^2 \, \big{|} {\bf y}, {\bf \theta}^{(t)} \right ] $ can be efficiently computed. To deal with process (\ref{Cross}) we proceed as follows
\begin{eqnarray*} 
 {\textrm E}\left [ l_j  \big{|} {\bf y}, {\bf \theta}^{(t)}  \right ]  & = &  
\int d{\bf l} \, l_j \, \rho({\bf l} \, \big{|}  {\bf y}, {\bf \theta}^{(t)})
\end{eqnarray*}
\begin{eqnarray} 
 & = & \frac{\int d{\bf l} d{\bf \beta} \,\, l_j \, \rho({\bf y}, {\bf l}, {\bf \theta}^{(t)}, {\bf \beta})}{\int d{\bf l} d{\bf \beta} \,\, \rho({\bf y}, {\bf l}, {\bf \theta}^{(t)}, {\bf \beta})} \nonumber \\
& = & \frac{\int d{\bf l} d{\bf \beta} \,\, l_j \, \rho({\bf y} \big{|} {\bf l}, {\bf \theta}^{(t)}) \, \rho({\bf \theta}^{(t)} \big{|} {\bf \beta}) \, \rho( {\bf \beta} \big{|} {\bf l} ) \, \rho({\bf l}) }{\int d{\bf l} d{\bf \beta} \, \rho({\bf y} \big{|} {\bf l}, {\bf \theta}^{(t)}) \, \rho({\bf \theta}^{(t)} \big{|} {\bf \beta}) \, \rho( {\bf \beta} \big{|} {\bf l} ) \, \rho({\bf l})} \nonumber 
 \end{eqnarray}
and making use of the same Laplacian\footnote{we are keeping only linear correlations in the posterior distribution of ${\bf l}$, but strictly speaking this is not a Laplace approximation} approximation as above, we find
\begin{eqnarray}
& \approx & \frac{\int d{\bf l} d{\bf \beta} \,\, l_j \, \rho({\bf y} \big{|} {\bf l}, {\bf \theta}^{(t)}) \, \rho({\bf \theta}^{(t)} \big{|} {\bf \beta}) \, \rho( {\bf \beta} \big{|} {\bf l}^{(t)} ) \, \rho({\bf l}) }{\int d{\bf l} d{\bf \beta} \, \rho({\bf y} \big{|} {\bf l}, {\bf \theta}^{(t)}) \, \rho({\bf \theta}^{(t)} \big{|} {\bf \beta}) \, \rho( {\bf \beta} \big{|} {\bf l}^{(t)} ) \, \rho({\bf l})} \nonumber \\
& = & \frac{\int d{\bf \beta} \, \rho({\bf \theta}^{(t)} \big{|} {\bf \beta}) \, \rho( {\bf \beta} \big{|} {\bf l}^{(t)} ) \, \, \int d{\bf l} \,\, l_j \, \rho({\bf y} \big{|} {\bf l}, {\bf \theta}^{(t)}) \, \rho({\bf l}) }{\int d{\bf \beta} \,\, \rho({\bf \theta}^{(t)} \big{|} {\bf \beta}) \, \rho( {\bf \beta} \big{|} {\bf l}^{(t)} ) \, \, \int d{\bf l} \, \rho({\bf y} \big{|} {\bf l}, {\bf \theta}^{(t)}) \, \rho({\bf l}) } \nonumber \\
& = & \frac{\int d{\bf l} \,\, l_j \, \rho({\bf y} \big{|} {\bf l}, {\bf \theta}^{(t)}) \, \rho({\bf l}) }{ \int d{\bf l} \, \rho({\bf y} \big{|} {\bf l}, {\bf \theta}^{(t)}) \, \rho({\bf l}) } \nonumber \\
& = & {\textrm E}^{*} \left [ l_j  \big{|} {\bf y}, {\bf \theta}^{(t)} \right ]  \nonumber
\end{eqnarray}
Similarly, we have
\begin{eqnarray}
{\textrm E}\left [ l_j^2  \big{|} {\bf y}, {\bf \theta}^{(t)} \right ] & \approx & {\textrm E}^{*} \left [ l_j^2  \big{|} {\bf y}, {\bf \theta}^{(t)} \right ]  \nonumber
\end{eqnarray}


\section*{Acknowledgment}

The author wishes to thank Natalia Barsheshet for her always original and inspiring insights.

\ifCLASSOPTIONcaptionsoff
  \newpage
\fi


\begin{thebibliography}{17}

\bibitem{Shannon:1948}
C.E. Shannon, \emph{A mathematical theory of communication},
Bell System Technical Journal, vol. 27, pp. 379-423 and 623-656, July and October, 1948.

\bibitem{Cok:1986}
D.R. Cok, \emph{Signal processing method and apparatus for producing interpolated chrominance values in a sampled color image
signal},
 U.S. patent 4,642,678, 1986

\bibitem{Adams:1996}
J.E. Adams and J.F. Hamilton Jr, \emph{Adaptive color plane interpolation in single color electronic camera},
 U.S. patent 5,506,619, 1996

\bibitem{Kimmel:1999}
R. Kimmel, \emph{Demosaicing: image reconstruction from color CCD samples},
 IEEE Trans. Image Processing, vol. 8, no. 9, 1221-1228, 1999

\bibitem{Wu:1997}
X. Wu, W.K. Choi, and P. Bao, \emph{Color restoration from digital camera by pattern matching},
 Proc.SPIE, vol.3018, pp.12-17, 1997

\bibitem{Lu:2003}
W. Lu and Y.-P. Tan, \emph{Color filter array demosaicking: New method and performance measures},
 IEEE Trans. Image Processing, vol. 12, no. 10,
pp. 1194-1210, 2003

\bibitem{Hibbard:1995}
R.H. Hibbard, \emph{Apparatus and method for adaptively interpolating a full color image utilizing
luminance gradients},
 U.S. patent 5,382,976 1995

\bibitem{Laroche:1994}
C.A. Laroche and M.A. Prescott, \emph{Apparatus and method for adaptively interpolating a full color image utilizing
chrominance gradients},
 US Patent, 5373322, 1994

\bibitem{Hamilton:1997}
J.F. Hamilton and J.E. Adams, \emph{Adaptive color plan interpolation in single sensor color electronic camera},
 US Patent,
5,629,734, 1997

\bibitem{Chang:1999}
E. Chang, S. Cheung, and D. Y. Pan, \emph{Color filter array recovery using a threshold-based variable number of gradients},
in Sensors, Cameras, and Applications for Digital Photography, Proc. SPIE, vol. 3650, pp. 36-43, 1999

\bibitem{Li:2008}
X. Li, B. Gunturk, L. Zhang, \emph{Image demosaicing: a systematic survey},
in Visual Communications and Image Processing, Proc.SPIE, vol. 6822, pp. 68221J-68221J-15, 2008

\bibitem{Bousquet:2004}
O. Bousquet, S. Boucheron, G. Lugosi \emph{Introduction to Statistical Learning Theory},
Advanced Lectures on Machine Learning, Springer, pp. 169-207, 2004

\bibitem{Brainard:1994}
D.H. Brainard, \emph{Bayesian method for reconstructing color images from trichromatic samples},
Proceedings of the IS\&T 47th Annual Meeting, Rochester, NY, 375-380, 1994

\bibitem{Brainard:1997}
D.H. Brainard and W.T. Freeman, \emph{Bayesian color constancy}, 
Journal of the Optical Society of America A, 14, 1393-1411, 1997
 
\bibitem{Fukunaga:1990}
K. Fukunaga, \emph{Introduction to Statistical Pattern Recognition},
second ed., Academic Press, New York, 1990

\bibitem{Shiryaev:1996}
A.N. Shiryaev, \emph{Probability},
Graduate Texts in Mathematics, Springer, New York, 1996 

\bibitem{Taubman:2000}
D. Taubman, \emph{Generalized Wiener reconstruction of images from colour sensor data using a scale invariant prior}, 
in IEEE Int. Conf. on Image Processing, vol. 3, pp. 801-804, 2000

\bibitem{Roth:2005}
S. Roth and M.J. Black, \emph{Fields of Experts: A Framework for Learning Image Priors},
IEEE Conference on Computer Vision and Pattern Recognition, vol. 2, pp 860-867, 2005

\bibitem{Mairal:2008}
J. Mairal, M. Elad, and G. Sapiro, \emph{Sparse Representation for Color Image Restoration}, 
IEEE Trans. on Image Processing, Vol. 17, No. 1, pages 53-69, 2008

\bibitem{Hel-Or:2002}
Y. Hel-Or and D. Keren, \emph{Demosaicing of Color Images Using Steerable Wavelets},
HP Labs Israel, Tech. Rep. HPL-2002-206R1 20020830, 2002

\bibitem{Mukherjee:2001}
J. Mukherjee, R. Parthasarathi, and S. Goyal, \emph{Markov random field processing for color demosaicing},
Pattern Recognition Letters, vol. 22, no. 3-4, pp. 339-351, 2001

\bibitem{Geman:1984}
S. Geman and D. Geman, \emph{Stochastic Relaxation, Gibbs Distributions, and the Bayesian Restoration of Images},
IEEE Transactions on Pattern Analysis and Machine Intelligence, vol. 6, no. 6, pp. 721-741, 1984

\bibitem{Kschischang:2001}
F.R. Kschischang, B.J. Frey and H-A. Loeliger \emph{Factor graphs and the sun-product algorithm} ,
IEEE Trans Information Theory, vol. 47 , pp. 498-519, 2001.

\bibitem{Yedidia:2005}
J.S. Yedidia, W.T. Freeman and Y. Weiss, \emph{Constructing free-energy approximations and generalized belief propagation algorithms} ,
IEEE Trans on Information Theory, vol. 51, pp. 2282-2312, 2005.

\bibitem{Dempster:1977}
A. Dempster, N. Laird and D. Rubin, \emph{Maximum Likelihood from incomplete data via the EM algorithm},
Journal of the Royal Statistical Society, Series B,39, pp. 1-38, 1977

\bibitem{Go:2000}
J. Go, K. Sohn, and C. Lee, \emph{Interpolation using neural networks for digital still cameras},
IEEE Trans. Consumer Electronics, 46(3):610-616, 2000

\bibitem{Kapah:2000}
O. Kapah and Y. Hel-Or, \emph{Demosaicing using artificial neural networks},
in Proc. SPIE, vol. 3962, pp.112-120, 2000

\bibitem{Ruderman:1994}
D.L. Ruderman and W. Bialek, \emph{Statistics of natural images: scaling in the woods},
 Phys. Rev. Lett, vol. 73, pp. 814-817, 1994

\bibitem{Kalman:1960}
R.E. Kalman, \emph{A new approach to linear filtering and prediction problems}, 
Journal of Basic Engineering, vol. 82, pp. 35-45, 1960

\bibitem{Simoncelli:1996}
E.P. Simoncelli and E.H. Adelson, \emph{Noise Removal via Bayesian Wavelet Coring} ,
Proc 3rd IEEE Int'l Conf on Image Proc, vol. 1 , pp. 379-382, 1996.

\bibitem{Nocedal:1995}
R.H. Byrd, P. Lu, J. Nocedal and C. Zhou, \emph{A limited memory algorithm for bound constrained optimization},
SIAM Journal on Scientific and Statistical Computing, vol. 16, pp. 1190-1208, 1995.

\bibitem{Gallager:1963}
R.G. Gallager, \emph{Low-density parity check codes},
MIT Press, 1963

\bibitem{Sudderth:2003}
E.B. Sudderth, A. Ihler, W.T. Freeman and A. Wilsky, \emph{Nonparametric belief
propagation},
CVPR, vol. 1, pp. 605-612, 2003

\bibitem{Wu:1983}
C.F.J. Wu, \emph{On the convergence properties of the EM algorithm},
Annals of statistics,  vol. 11, pp. 95-103, 1985

\bibitem{Lukac:2005}
R. Lukac and K.N. Plataniotis, \emph{Data-Adaptive Filters for Demosaicking: A Framework},
IEEE Trans. on Consumer Electronics, vol. 51, no. 2, pp. 560-570, 2005

\bibitem{Breiman:1984}
L. Breiman, J.H. Friedman, R.A. Olshen, and C.J. Stone, \emph{Classification
and Regression Trees},
Wadsworth, Belmont, 1984

\bibitem{UC-SIPI}
{\em http://r0k.us/graphics/kodak}

\bibitem{Li:2005}
X. Li, \emph{Demosaicing by successive approximation},
IEEE Trans. on Image Processing,
vol. 14, no. 3, pp. 370-379, 2005

\bibitem{SCIELAB}
Xuemei Zhang,
{\em http://white.stanford.edu/~brian/scielab/scielab.html}

\bibitem{Longere:2002}
P. Longere, X. Zhang, P. B. Delahunt, and D. H. Brainard, \emph{Perceptual assessment of demosaicing algorithm performance},
Proc. IEEE, vol. 90, no. 1, pp. 123132, 2002

%\bibitem{HelOr:2002}
%Y. Hel-Or and D. Keren, \emph{Image demosaicing method using directional smoothing} ,
%U.S. Patent 6404918, 2002.

%\bibitem{Keren:1999}
%D. Keren and M. Osadchy, \emph{Restoring subsampled color images},
% Machine vision and applications, vol. 11, pp. 197-202, 1999.

%\bibitem{Paliy:2007}
%D. Paliy, V. Katkovnik, R. Bilcu, S. Alenius, and K. Egiazarian, \emph{Spatially adaptive
%color filter array interpolation for noiseless and noisy data},
%International Journal of Imaging Systems and Technology, vol. 17, no. 3, 
%pp. 105122, 2007

\end{thebibliography}
\end{document}